%% file: vaetestgen2.tex
\documentclass[conference]{IEEEtran}
\IEEEoverridecommandlockouts

\newcommand{\etal}{{\it et~al}.}

\usepackage{cite}
\usepackage{algorithm,algorithmic}
\usepackage{subfig}
\usepackage{multirow}
\usepackage{booktabs}
\usepackage{amsmath,amssymb,amsfonts}
\usepackage{graphicx}
\usepackage{comment}
\usepackage{textcomp}
\usepackage{xcolor}
\usepackage[bookmarks=false]{hyperref}
\usepackage{todonotes}

\usepackage[super]{nth}
\usepackage{esvect}

\begin{document}

\title{\makebox[\linewidth]{\parbox{\dimexpr\textwidth+2cm\relax}{\centering Manifold-based Test Generation for Image Classifiers}}}


\author{
\IEEEauthorblockN{Taejoon Byun}
\IEEEauthorblockA{
    \textit{Computer Science \& Eng.} \\
    \textit{University of Minnesota} \\
    Minneapolis, MN, USA \\
    taejoon@umn.edu
    }
\and
\IEEEauthorblockN{Abhishek Vijayakumar}
\IEEEauthorblockA{
    \textit{School of Computer Science} \\
    \textit{Carnegie Mellon University}\\
    Pittsburgh, PA, USA\\
    vijayakumar.abhishek@gmail.com
    }
\and
\IEEEauthorblockN{Sanjai Rayadurgam}
\IEEEauthorblockA{
    \textit{Computer Science \& Eng.} \\
    \textit{University of Minnesota} \\
    Minneapolis, MN, USA\\
    rsanjai@umn.edu
    }
\and
\IEEEauthorblockN{Darren Cofer}
\IEEEauthorblockA{
    \textit{Advanced Technology Center} \\
    \textit{Collins Aerospace}\\
    Bloomington, MN, USA\\
    darren.cofer@collins.com
    }
}


\maketitle

\begin{abstract}
Neural networks used for image classification tasks in critical applications must be tested with sufficient realistic data to assure their correctness.
%
%
This raises two challenges: first, an adequate subset of the data points must be carefully chosen to inspire confidence, and second, the implicit requirements must be meaningfully extrapolated to data points beyond those in the explicit training set.
This paper proposes a novel framework to address these challenges.
Our approach is based on the premise that patterns in a large input data space can be effectively captured in a smaller manifold space, from which similar yet novel test cases---both the input and the label---can be sampled and generated.
A variant of Conditional Variational Autoencoder (CVAE) is used for capturing this manifold with a generative function, and a search technique is applied on this manifold space to efficiently find fault-revealing inputs.
Experiments show that this approach enables generation of thousands of realistic yet fault-revealing test cases efficiently even for well-trained models.
\end{abstract}


\begin{IEEEkeywords}
machine learning testing, test generation, neural networks, variational autoencoder
\end{IEEEkeywords}



\section{Introduction}
\input{sections/intro.tex}

\section{Preliminaries}
\input{sections/background.tex}

\section{Manifold-based Test Generation}
\input{sections/testgen.tex}

\section{Experiment}
\input{sections/experiment.tex}

\section{Result}
\input{sections/result.tex}


\section{Related Work}
\input{sections/related.tex}

\section{Conclusion and Future Work}
\input{sections/conclusion.tex}

\section*{Acknowledgement}
This work was supported by AFRL and DARPA under contract
FA8750-18-C-0099.

\bibliographystyle{IEEEtran}
\bibliography{vaetestgen2}

\end{document}

%% file: sections/intro.tex
The increasing use of machine-learning components such as deep neural networks in safety-critical applications is driving the focus of software engineering researchers toward important questions of verification and validation of such systems~\cite{zhang2019machine}.
%
%
As several works in adversarial input generation~\cite{zhang2019machine} have shown the susceptibility of typical image classifiers to input transformation attacks, there is an unmet but urgent need for principled approaches to testing of such systems.

An effective testing regime must adequately exercise the system under test to inspire confidence that any discrepancy between the system's implemented---or learned---behavior and its specified requirements is exposed.
Test generation can effectively address this problem if it can automatically generate 1) realistic, 2) complete with respect to the requirement, and 3) fault-revealing test cases.
For neural network image classifiers, however, each of these goals is challenging to achieve.
First, realistic images are hard to be generated automatically because they lie in a high dimension.
Second, it is not easy to capture the requirement nor determine the completeness of a test suite with respect to it, as the requirement is implicit in the training data.
Third, a test input has to be labeled with a correct oracle which is labor intensive if not automated.
Existing works attempt to tackle some of these problems, mostly by utilizing metamorphic relation and/or using generative models~\cite{zhu2019datamorphic}.
DeepTest~\cite{tian2017deeptest} and DeepRoad~\cite{zhang2018deeproad}, for instance, can generate realistic and fault-finding test cases by hinging upon oracle-preserving metamorphic transformations, and by utilizing image translations enabled by image filters or generative adversarial networks.
Although these techniques can be effective, they cannot generate less dramatic but arguably more important test cases---normal and realistic cases that look just like any of the training data but trigger failures.

This paper proposes a novel approach that can effectively address the above-mentioned challenges and complement metamorphic test generation techniques.
We capture a domain model of the dataset using a variant of conditional variational autoencoder (CVAE), an unsupervised learning technique which can learn a conditioned manifold---compact representations of the dataset per label---along with an encoder and a decoder that can map the dataset to and from the manifold.
Once a VAE is trained, new test cases can be sampled from this manifold and mapped to the input dimension using the decoder.
These test inputs are likely realistic in-distribution images, as a VAE is optimized towards producing such images with high probability.
These inputs can also be novel to an extent that a VAE can interpolate among existing data points~\cite{berthelot2018understanding}, allowing new problems to be discovered using these test inputs.
The key idea of this paper, then, is to apply search-based test generation~\cite{mcminn2011search} on the manifold space so that novel, interesting, and fault-revealing test cases are generated.
A fitness function is defined such that the uncertainty of the model under test is maximized, the rationale being that high uncertainty inputs are more likely to trigger faults~\cite{byun2019input}.

We evaluated the proposed approach with three popular image classification tasks---MNIST, Fashion MNIST, and CIFAR---and one in-house image classification task---TaxiNet.
For a set of well-trained models for the respective tasks, we study the fault-finding effectiveness of the generated test cases, and assess the realism of the test cases both qualitatively and quantitatively.
The results show that the proposed approach can indeed generate realistic yet fault-revealing test cases effectively with a minimal human intervention in the process.

%% file: sections/background.tex
A \emph{manifold} is formally a topological space that is locally Euclidean (e.g. the surface of the Earth).
%
The basic premise in manifold learning is  that real world data $X$ presented in high-dimensional spaces $R^{d_X}$ are expected to concentrate in the vicinity of a manifold $M$ of a much lower dimension $d_M$ embedded in $R^{d_X}$~\cite{bengio2013representation}.
In other words, high dimensional data---such as image---can be explained with a number of factors that is much smaller than the dimensionality of the input space.   
%
%
Manifold learning tries to capture such mapping so that a complex dataset can be encoded into a meaningful representation in a smaller dimension, serving several purposes such as data compression and visualization~\cite{cayton2005algorithms}.
We have developed our techniques by leveraging variational autoencoders (VAE) as they provide unique capabilities for synthesizing new inputs from the manifold.

%

%

\subsection{Variational Autoencoder}

VAE is a latent-variable generative model capable of producing outputs similar to inputs by determining a latent-variable space $Z$ and associated probability density function (PDF) $P(z)$.
%
The goal of a latent-variable model is to make sure that, for every datapoint $x$ in a given dataset $X$, there is one or more setting of the latent variables $z$ in a space $Z$ which causes the model to generate $\hat{x}$ that is very similar to $x$.
This goal is achieved by optimizing $\theta$ for a deterministic function $f: Z \times \Theta \to X$ such that the random variable $f(z; \theta)$ produces outputs similar to  $x \in X$ when $z$ is sampled from $P(z)$.
In other words, we maximize the likelihood of producing $X$ when $X$ is conditioned by $Z$: $P(X) = \int P(X | z; \theta) P(z) dz$; here, a PDF $P(X|z; \theta)$ replaces $f(z; \theta)$.
VAE does not assume a specific distribution for $P(z)$, but rather assumes that any probability distribution in the space $Z$ can be represented by applying a sufficiently complicated function $f_\theta$ to a set of normally distributed variables $z$.
With a set of decoder parameters $\theta$, the probabilistic decoder of a VAE is given by:
\begin{equation}
    P_{\theta} (x|z) = \mathcal{N} (x|f_{\mu_x}(z;\theta), \gamma I)
\end{equation}
where $\gamma$ is a tuneable scalar hyperparameter---which is typically set as 1 to represent multivariate unit Gaussian distribution---and $I$ is the identity matrix.
We set $\gamma$ as a trainable parameter, as a high $\gamma$ is proven to be responsible for blurry images generated by VAEs, which was often considered as a practical limitation of VAEs~\cite{dai2019diagnosing}.

For modeling the unknown PDF of latent variables $P(z|x)$ from which to run the decoder $P_\theta (x|z)$, we need a new PDF $Q(z|X)$ which can take an $x$ and return a distribution over $z$ that are likely to produce $x$.
This $Q(z|x)$ is called probabilistic encoder, which is given by:
\begin{equation}
    Q_{\phi} (z|x) = \mathcal{N} (z | g_{\mu_z} (x; \phi), g_{{\sigma_z}^2}(x; \phi))
\end{equation}
where $\phi$ is a set of encoder parameters and $g$ is an encoder function approximated by a deep neural network.
$g$ is designed to produce two outputs $g_{\mu_z}$ and $g_{{\sigma_z}^2}$, which are mean and variance of the encoded $z$.
In other words, $g$ encodes each $x \in X$ as a distribution, where the mean $g_{\mu_z}$ has the highest probability of being reconstructed to $x$.

As $P(z|x)$ was assumed as multivariate Gaussian, the posterior distribution $Q_\phi (z|x)$ shall {\it match} the $P(z|x)$ so that we can relate $P(x)$ to $\mathbb{E}_{z \sim Q} P(x|z)$, or the expected value of generated input $x$ given a latent variable $z$ when $z$ is sampled from the space encoded by encoder PDF $Q$.
This is achieved by optimizing the following VAE loss function:
\begin{multline}
    \mathcal{L}(\theta,\phi) = \int_{\mathcal{X}} - \mathbb{E}_{Q_\phi (z|x)}
    \lbrack \log P_\theta (x|z) \rbrack \\
    + \mathbb{KL} \lbrack Q_\phi (z|x) || P(z) \rbrack \mu_{gt} (dx)
    \label{eq:vae}
\end{multline}
where $\mu_{gt}(dx)$ is the ground-truth probability mass of a $dx$ on $X$, which leads to $\int_{X} \mu_{gt}(dx) = 1$.
%
%
The term $- \mathbb{E}_{q_\phi (z|x)} \lbrack \log p_\theta (x|z) \rbrack$ is the reconstruction cost, which penalizes poor reconstruction inputs in the input dataset.
The term $\mathbb{KL} \lbrack q_\phi (z|x) || P(z) \rbrack$ is the Kullback-Leibler divergence between the encoder distribution and the prior distribution, which penalizes deviations from the distribution $P(z)$.
~\cite{dai2019diagnosing}.
We defer more curious readers to a tutorial on VAE~\cite{doersch2016tutorial}.

\subsection{A VAE example}

\begin{figure}[h]
    \centering
    \includegraphics[width=0.85\columnwidth]{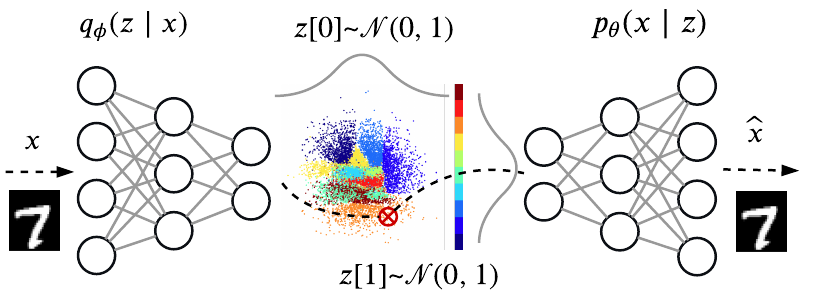}
    \caption{Variational Autoencoder}
    \label{fig:vae}
\end{figure}

Figure~\ref{fig:vae} illustrates the structure and the operation of a VAE with $\kappa = 2$ as the size of the latent dimension, and $\gamma = 1$.
VAE encodes $x$ as $z \sim \mathcal{N}(0, I_2)$, where $I_2$ is a $2 \times 2$ identity matrix, forming a circle-like mappings to the two-dimensional plan.
As 99.73\% of the datapoints fall inside the range $[-3\sigma, 3\sigma]$, the majority of the datapoints fall inside a circle of radius $3\sigma = 3$.
The datapoints mapped in the plane shows that digits in the same class---color-coded from 0 (dark blue) to 9 (dark brown)---cluster together, illustrating that digits that {\it look} similar are encoded to be close to each other in the latent space.
From the areas where different colors are mixed together, such as where mint-colored points representing digit 4 and dark-brown-colored points representing 9 are mixed in an adjacent space, we can infer that many fours and nines look similar to each other.
If we sample new points from this subspace, the generated image may look somewhat like 4 and 9 at the same time.

\subsection{Conditional VAE}

A vanilla VAE can generate images but not labels.
Thus, it may be useful for test input generation, but without the labels, much time has to be spent for assigning labels to solve the oracle problem.
When implementing VAE, note that the encoder $q_{\phi} (z|x)$ is conditioned solely on the inputs $x$, and similarly, the decoder $p_{\theta} (x|z)$ models $x$ solely based on the latent-variable vector $z$.
Conditional VAE (CVAE) implements a conditional variable $c$ in both the encoder and decoder~\cite{larsen2015autoencoding}.
This yields the new loss function
\begin{multline} 
  \mathcal{L}(\theta,\phi) = \int_{\mathcal{X}} - \mathbb{E}_{q_\phi (z|x)}
  \lbrack \log p_\theta (x|z,c) \rbrack \\
  + \mathbb{KL} \lbrack q_\phi (z|x,c) || P(z|c) \rbrack \mu_{gt} (dx)
  \label{eq:cvae}
\end{multline}
Note that $P(z)$ is now distributed under $P(z|c)$, a conditional distribution on $c$.
Both the encoder and decoder are conditioned on $c$ as well, which gives a specific distribution $P(z|c)$ for each class $c$.
When training a CVAE for a classification task, we choose the values of $c$ to be the class labels of the dataset.
By sampling $z$ from $P(z|c)$, we significantly increase the probability of obtaining a latent-variable vector $z_0$ such that $p_\theta (x, z_0,c)$ is a valid image of class $c$~\cite{sohn2015learning}.

%

\subsection{Two-stage VAE}

One drawback to the vanilla VAE is the inability to accurately reproduce the distribution of ground-truth data $P_{gt} (x)$, even with perfect reconstruction loss.
Although the prior distribution $P(z)$ is a Gaussian, the encoder projects the ground-truth distribution as $Q(z)$ which is not necessarily Gaussian at the global optimum.
As a result, when generating synthetic outputs using a vanilla VAE, the distribution of sampled latent vectors from $P(z)$ does not match the distribution of the ground-truth latent vectors $Q(z)$, and the ground-truth distribution $X$ is not accurately reproduced.

In order to address this problem, Dai and Wipf introduced two-stage VAE~\cite{dai2019diagnosing}, which makes use of a second-stage latent space $U$ and associated density function $P(u)$.
Simply speaking, after training the (first-stage) VAE with Equation~\ref{eq:vae} to generate $\hat{x}$, another (second-stage) VAE is trained with the following second-stage loss function:
\begin{multline}
    \mathcal{L}(\theta',\phi') = \int_{Z} - \mathbb{E}_{Q'_{\phi'} (u|z)}
    \lbrack \log P'_{\theta'} (z|u) \rbrack \\
    + \mathbb{KL} \lbrack Q'_{\phi'} (u|z) || P'(u) \rbrack \mu_{gt} (dz)
    \label{eq:vae2}
\end{multline}
Note that it is in the same form as Equation~\ref{eq:vae}, with the $\theta$, $\phi$, $Q$, $P$, $z$, and $x$ replaced with $\theta'$, $\phi'$, $Q'$, $P'$, $u$, and $z$, respectively.
In other words, this second-stage VAE is trained with $Z$ as the input dataset, and learns $u$ as latent variables with which to encode $Z$.
This second-stage VAE resolves the discrepancy between the prior distribution $P(z)$ and posterior distribution $Q(z|x)$ by introducing a second-stage latent distribution $P'(u)$ from which to sample new inputs.
The second-stage latent distribution is proved to fit better to Gaussian prior, such that when new inputs are sampled from $u ~\mathcal{N}(0, I_\kappa)$ with $\kappa$ being the size of the latent dimension, the reconstructed $\hat{z}$ lies in the ground truth distribution $Q(z)$.
The $\hat{z}$ is then fed to the first-stage decoder, which generates $\hat{x}$ with a high $P(x)$.

\subsection{Two-stage Conditional VAE}

We created a novel conditional two-stage VAE by nesting two CVAEs and training them just like Two-Stage VAEs.
Essentially, it is a CVAE with improved sampling quality.

%% file: sections/testgen.tex
The goal of testing an image classifier is to find {\it faults} in a model which cause discordances between existing conditions and required conditions~\cite{zhang2019machine}.
For a given image classification model $M$, a test case $(x, y)$, an (input, expected output) pair, is fault-revealing if 1) $x$ is in-distribution---i.e. satisfies the assumption of the model---and 2) the output of the model $M(x)$ is different from $y$.
%
As with training data, such test cases can be collected and manually labeled, but doing so can become very expensive, especially as the accuracy of the model under nears perfection.
%

%
%
%
%

Automated test generation attempts to solve this problem by synthesizing test cases---or pairs of a test input and the expected output---that are likely to reveal faults in the model~\cite{pei2017deepxplore,zhang2018deeproad,tian2017deeptest}.
If one can reliably generate realistic in-distribution images that are revealing faults, the testing process can be much accelerated since those synthesized tests can serve as counter-examples, highlighting where the weakness of the model lies. 
However, generating realistic yet fault-revealing images is not a trivial task because images reside in a very high dimension---even for a very small gray-scale image of $28 \times 28$ pixels, there can be $256^{28 \times 28}$ possibilities.
When a random testing is applied on the input space, the probability of sampling a realistic input would be close to zero.
Search-based testing cannot be applied neither, since the search space is too huge to be handled efficiently.
%
This section introduces our technique for synthesizing realistic test inputs along with expected outputs automatically.

\subsection{The Approach}

\begin{figure}[h]
    \centering
      \includegraphics[width=0.9\columnwidth]{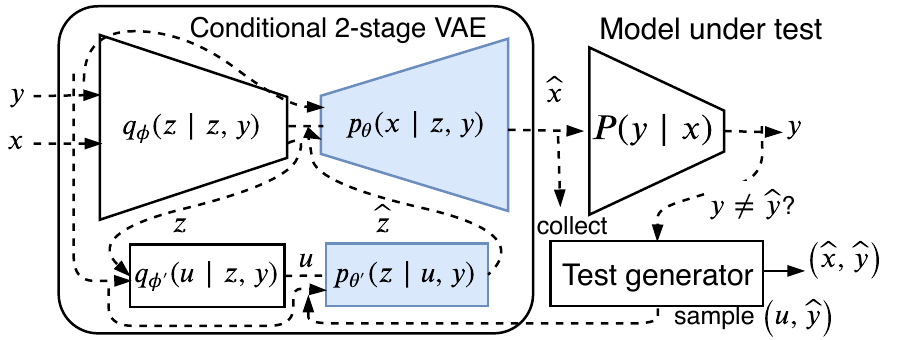}
    \caption{The Manifold-based Test Generation Framework}
    \label{fig:structure}
\end{figure}

The key idea of our approach is to apply search-based test generation~\cite{mcminn2011search} on the manifold space.
As a technique for learning a manifold, and also for obtaining a generative model, we train a Conditional Variational Autoencoder (CVAE).
To facilitate the search, we introduce a fitness function which can evaluate the relative value of a sampled data point in the latent space.
The fitness is also judged upon the output and the {\it sentiment}~\cite{byun2019input} of the model under test; hence, the overall framework of our approach as shown in Figure~\ref{fig:structure}.

%
%

Figure~\ref{fig:structure} shows the framework of our manifold-based test generation.
First of all, a two-stage VAE---which is the state-of-the-art VAE at the time of writing this paper---is trained one after the other as suggested by Dai and Wipf~\cite{dai2019diagnosing}.
The first VAE is trained to reconstruct the original input image $x$ as $\hat{x}$ while learning the first-stage manifold $P(z|x)$.
The second VAE is trained to reconstruct the first-stage encoding $z$ as $\hat{z}$ while learning the second-stage manifold $P'(u|z)$.
For both VAEs, the labels of the images are given to the encoder and the decoder as condition $c$ during training, so that we can condition the generated test inputs with desired outputs.
Once the VAEs are trained, the encoders are discarded and only the decoders are used for generating new images.
A new sample $\hat{x}$ can be synthesized by feeding in a choice of vector $u$ with a condition $c$ to the second-stage decoder, and in turn $\hat{u}$ and $c$ to the subsequent first-stage decoder.
In other words, the generated input $\hat{x}$ is solely dependent on the choice of a latent vector $u$ and a conditioning label $c$.
By claiming a control over this second-stage latent space $U$, and by introducing a fitness function, a fitness landscape can be drawn over this space, from which we can generate test inputs of our liking---fault-revealing, {\it realistic}, and {\it interesting}.

The detailed procedure of manifold-based test generation is described in Algorithm~\ref{alg}.
For each test case to generate, a label $\hat{y}$ is sampled first to condition the decoders.
Second, an arbitrary second-stage latent vector $u$ is sampled with our sampling method which will be explained later.
Third, the selected pair $(u, \hat{y})$ is passed to the decoders, and new test input $\hat{x}$ is synthesized.
To see if the new test case$(\hat{x}, \hat{y})$ reveals a fault in the model, we obtain the prediction of the model for $\hat{x}$ and check if the predicted output matches the label.
If a mismatch is found, the pair $(u, \hat{y})$ which generated new input $\hat{x}$ is compared against all the previously collected latent vectors to find if a new $u$ had been sampled before.
We assume that a distinct $u$ leads to a distinct $\hat{x}$.
The algorithm terminates once all the $N$ test cases are generated.

 \begin{algorithm}[h]
 \caption{Manifold-based test case generation algorithm}
 \label{alg}
 \begin{algorithmic}[1]
 \small
 \renewcommand{\algorithmicrequire}{\textbf{Input:}}
 \renewcommand{\algorithmicensure}{\textbf{Output:}}
 \REQUIRE $m$: model under test, $v_1$: \nth{1}-stage VAE, $v_2$: \nth{2}-stage VAE, $N$: number of tests to generate, $k$: distance threshold between latent vectors, $n$: number of optimization iterations
 \ENSURE  $T$: a set of test cases, where $T_i = (\hat{x}_i, \hat{y}_i)$
 \STATE $U \leftarrow \phi$; $T \leftarrow \phi$; $is\_duplicate \leftarrow False$
 \WHILE {$|T| \le N$}
    \STATE $\hat{y} \leftarrow$ {\it random integer} $\in [0, m.num\_classes)$
    \STATE $u \leftarrow${\bf sample($m$, $v_1$, $v_2$)}
    \STATE $\hat{z} \gets v_2$.decode($u$, $\hat{y}$)
    \STATE $\hat{x} \gets v_1$.decode($\hat{z}$, $\hat{y}$) 
    \STATE $y \gets$ m.predict($\hat{x}$) 
\\ \textit{$\hat{x}$ is fault-revealing when $y \ne \hat{y}$}
    \IF {$y \ne \hat{y}$}
    
        \FORALL{$\{(\mu_i, \hat{y}_i) | \hat{y}_i = \hat{y} \wedge (\mu_i, \hat{y}_i) \in U \}$}
            \IF {$|\mu_i - u|_{l_2} < k$}
                \STATE $is\_duplicate \leftarrow True$; break
            \ENDIF
        \ENDFOR
        \IF {!$is\_duplicate$}
            \STATE $T \leftarrow T \cup \{(\hat{x}, \hat{y})\}$; $U \leftarrow U \cup \{(u, y)\}$
            \STATE $is\_duplicate \leftarrow False$
        \ENDIF
    \ENDIF 
 \ENDWHILE
 \RETURN $T$ 
 \end{algorithmic} 
 \end{algorithm}

For sampling a latent vector $u$ from the $\kappa$-dimensional latent space (line 4 in Algorithm~\ref{alg}), we introduce two approaches---1) random sampling, and 2) optimization.
In the first approach, a latent vector $u$ is sampled repeatedly from the posterior distribution $\mathcal{N}_\kappa(0, I_\kappa)$ as in a typical use-case of the VAE decoder.
%
%
%
The second approach is to apply optimization with a fitness function which quantifies the merit of each sample point, and actively seek for a desirable sample.
With the optimization, a search metaheuristic can be applied to more carefully find optimal points.
%
%

%
%

%
%

%
%
%

\subsection{Search-based Optimization on Manifold}
\label{sec:fitness}

Once a desired amount of fault-revealing inputs are generated, one can optionally apply a fitness function, prioritize the test cases, and select the ones that yield high scores.
One of such fitness function can be defined in terms of Bayesian uncertainty~\cite{byun2019input}; the rationale behind preferring high-uncertainty inputs is that high uncertainty may indicate how {\it unfamiliar} a datum is to the model.
 
The first objective of our search is to find a $u$ which generates $\hat{x}$ that is most likely to show discordance between $y$ and $\hat{y}$.
We assume that this discordance can be captured as model's uncertainty $\sigma$, which can be approximately captured by enabling test-time dropout and performing Monte-Carlo simulation~\cite{gal2016dropout}.
This objective is formulated as 
\begin{equation}
    {\mathit o}_1 (u) = \frac{e^{\sigma(u)} - 1}{e^{\sigma(u)} + 1}
    \label{eq:opt}
\end{equation}
where $\sigma(u)$ is obtained by decoding $u$ to $\hat{x}$ and computing the uncertainty of the target model as in~\cite{byun2019input}.
The function $f(x) = (e^x - 1) / (e^x + 1)$ is a sigmoid-like normalization function that squashes any $\sigma \geq 0$ to $(0, 1)$.

When optimized towards uncertainty alone, however, we risk sampling an input of low likelihood in the latent distributions.
For example, when we train a VAE with latent dimension of $\kappa = 2$, the likelihood of sampling a point $(1, 1)$ from $\mathcal{N}_2(0, I_2)$ is about $0.0585$, while the likelihood of sampling $(3, 3)$ is merely $1.96 \times 10^{-5}$.
%
%
When a latent vector $u$ is sampled further away from populous regions where the encoding of the training data was concentrated, the decoder is less likely to produce high probability image $\hat{x}$, and $\hat{x}$ may look unrealistic, i.e. out of distribution.
To prevent this from happening, we introduce another optimization objective called {\it sample plausibility}, which is a slight modification to the probability density function of the unit normal distribution $f(x | \mathcal{N}(0, 1)) = (1 / \sqrt{2\pi}) e^{-0.5 x^2}$:
\begin{equation}
    {\mathit o}_2(u) = \sum_{i=0}^{\kappa}\frac{1}{e^{{u_i}^2}}
\end{equation}
The ${1}/{\sqrt{2\pi}}$ term is removed so that the maximum value of $o_2(u)$ is $1$ at $u$ is $\vv{0}$.
%
%
Values that deviate further from the mean are penalized so that $u$ is encouraged to stay close to the center in the space.

By combining the two objectives with corresponding weight hyper-parameters $w_1$ and $w_2$, we obtain the following cost function for joint-optimization:
\begin{equation}
    C(u) = w_1 {\mathit o}_1 (u) + w_2 {\mathit o}_2 (u)
\end{equation}
The weights are determined empirically such that both terms are equally optimized.

%


In our implementation, we used a stochastic optimization method known as particle swarm optimization (PSO)~\cite{kennedy2010particle} which simultaneously optimizes a population---or \textit{swarm}---of candidate solutions called \textit{particles}.
While particles have individual positions and velocities, their movements in the optimization space are determined by both their current behavior and the behavior of the swarm.
Particles are initialized at random points and initially sweep around the optimization space, eventually converging to local and global optima.

%% file: sections/experiment.tex
In the experiment, we evaluate the efficacy of manifold-based test generation.
We assess whether our technique can synthesize realistic test cases that reveal faults in the model under test, with the two approaches---random sampling and optimization.
The research questions are as follows:
%
\begin{itemize}
    \item \textbf{RQ1}: Can we generate realistic test inputs?
    \item \textbf{RQ2}: Can we generate fault-revealing test cases?
    \item \textbf{RQ3}: Is search better than random sampling?
\end{itemize}

\subsection{Model under test}

The experiments are performed with four different image classification tasks.
%
%
The first task is the popular MNIST hand-written digit classification~\cite{lecun1998mnist}.
The second task is fashion item image classification task, trained with Fashion-MNIST dataset~\cite{xiao2017fashion}.
Fashion MNIST is designed to be drop-in replacement for MNIST dataset, having the same image resolution and number of classes.
However, it is known to be more difficult than MNIST, with the state-of-the-art validation accuracy of 96.7\% with data augmentation.
%
The third one is CIFAR10~\cite{krizhevsky2009learning}, which is a ten-class image classification task with 50,000 training data.
Although the size of the images is small, the images are full-colored and complex, yet packed in a rather small 32 by 32 by 3 resolution.
The state-of-the-art accuracy without extra training data is 97.92\%~\cite{cai2018proxylessnas}.
The fourth task is TaxiNet, which is a dataset of runway images for autonomous taxiing task designed by our industry partner as a research prototype.
Since the original version of TaxiNet is designed to produce continuous values (a cross-track error from the runway center-line and relative heading deviation from the heading of the runway), we modified the design to produce a categorical output only in terms of the cross-track error---namely: far left, left, center, right, far right.

We trained one model for each task and achieved an accuracy close to the state-of-the-art.
For the architecture of the neural networks, we used some of the most popular (and standard) design techniques such as 2D convolution, max-pooling, batch normalization, and weight regularization.
We also introduced a dropout layer between the convolutions and fully-connected layer so that we can estimate the models' uncertainty using test-time dropout~\cite{gal2016dropout}.
The details of each model is described in Table~\ref{tbl:models}.
Note that we present the details for only two of the tasks, MNIST and Fashion MNIST, as our test generators could not reliably generate realistic inputs for the other tasks of higher dimension.

\begin{table}[h]
    \centering
    \caption{Models under test}
    \input{tables/models.tex}
    \label{tbl:models}
\end{table}

%
%

\subsection{Training Conditional Two-stage VAEs}

For the implementation of the Conditional Two-stage VAE, we revised the open-sourced TensorFlow code on GitHub~\cite{twostagevae} implemented by the authors of the Two-Stage VAE paper~\cite{dai2019diagnosing}.
For MNIST, Fashion MNIST, and CIFAR10, we used the model architecture inspired by InfoGAN; for TaxiNet, we used the ResNet-style architecture which makes use of residual connections inside the deep convolution layers.

\begin{table}[h]
    \centering
    \small
    \caption{Trained VAEs}
    \input{tables/vaes.tex}
    \label{tbl:vaes}
\end{table}

The success of a generative model can be measured by how faithfully it can produce a dataset that is similar to the training dataset, both in terms of the quality and the diversity.
Fr\'echet Inception Distance (FID) is a widely adopted measure that is shown to correlate well with human judgements~\cite{heusel2017gans}.
It measures the Fr\'echet distance between Gaussians fitted over the feature representations of the two datasets---training data and the generated data---where the feature representation is obtained by running a set of data through the Inception V3 network~\cite{szegedy2015going} and extracting 2048-dimensional feature representation in the pool3 bottleneck layer.
We report the two FID scores for each VAE---one for reconstructing the validation dataset and another for generating new dataset.
The former score shows how faithfully the model can encode and then decode the validation dataset---the set that it was not trained with---and generate the same dataset that it consumed---ideally, the reconstructed images shall look exactly the same to the original validation dataset when the VAE loss (Equation~\ref{eq:vae}) reached its global optimum.
The latter score shows the performance of the decoder alone, or the quality of the images generated by sampling from the prior imposed on the latent space.
Since our goal is to obtain a good generator, we tuned the VAE hyper-parameters for an optimal sampling FID score; lower score indicates a better quality.

The details of each VAE configuration and its FID score are shown in Table~\ref{tbl:vaes}.
The size of the latent dimension is a tunable hyper-parameter.
Ideally, an optimal VAE should be produced, when each dimension is fit to encode a latent feature with no redundancy.
This however, is only hypothetical, and we do not have control over the unsupervised process of learning the manifold.
We set the latent dimension size to 64 for all the tasks as suggested by Dai and Wipf~\cite{dai2019diagnosing}, and halved it to 32 if the sample FID score remained about the same.
Although it was argued in their paper that VAEs are trained to ignore superfluous latent dimensions, we observed that finding a smallest possible latent dimension size that retains the generation quality is a key in obtaining in-distribution inputs with our test generation algorithm.

For training the VAEs and performing the experiments, we used a Ubuntu 16.04 machine on Intel i5 CPU, 32GB DDR3 RAM, SSD, and a single NVIDIA GTX 1080-Ti GPU.


%

%% file: tables/models.tex
\footnotesize
\setlength\tabcolsep{4pt}
\begin{tabular}{@{}lllllll@{}}
\toprule
Task    & \begin{tabular}[c]{@{}l@{}}Image\\ Resolution\end{tabular} & \begin{tabular}[c]{@{}l@{}}\# Train\\ data\end{tabular} & \begin{tabular}[c]{@{}l@{}}\# Val.\\ data\end{tabular} & \begin{tabular}[c]{@{}l@{}}Train\\ acc.\end{tabular} & \begin{tabular}[c]{@{}l@{}}Val\\ acc.\end{tabular} & \begin{tabular}[c]{@{}l@{}}\# Train\\ params\end{tabular} \\ \midrule
MNIST   & $28 \times 28 \times 1$  & 60,000  & 10,000  & 99.38\%  & 99.15\%  & 594,922  \\
Fashion & $28 \times 28 \times 1$  & 60,000  & 10,000  & 99.86\%  & 92.71\%  & 858,026  \\
\bottomrule
\end{tabular}


%% file: tables/vaes.tex
\footnotesize
\begin{tabular}{@{}lrrrrr@{}}
\toprule
        & \multicolumn{1}{c}{\begin{tabular}[c]{@{}c@{}}Trainable\\ parameters\end{tabular}} & \multicolumn{1}{c}{\begin{tabular}[c]{@{}c@{}}Training\\ time\end{tabular}} & \multicolumn{1}{c}{\begin{tabular}[c]{@{}c@{}}Latent\\ dimension\end{tabular}} & \multicolumn{1}{c}{\begin{tabular}[c]{@{}c@{}}FID\\ recon.\end{tabular}} & \multicolumn{1}{c}{\begin{tabular}[c]{@{}c@{}}FID\\ sample\end{tabular}} \\ \midrule
MNIST   & 21,860,515 & 5h 21m & 32 & 6.64 & 8.80 \\
Fashion & 21,860,515 & 8h 19m & 32 & 17.97 & 23.86 \\
CIFAR10 & 26,074,053 & 6h 19m & 64 & 78.06 & 91.49 \\
TaxiNet & 41,294,597 & 13h 42m & 32 & 161.32 & 157.65 \\ \bottomrule
\end{tabular}

%% file: sections/result.tex
We ran our manifold-based test generator with the models under test in the loop, for generating fault-revealing test cases tailored for the models under test.
This section presents the result and address the three research questions.

\subsection{Can we generate realistic test inputs?}

\begin{figure*}[t]
    \centering
    \subfloat[MNIST]{
        \includegraphics[width=0.17\textwidth]{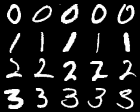}
        \label{fig:mnist_original}
    }
    \subfloat[Fashion MNIST]{
        \includegraphics[width=0.17\textwidth]{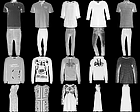}
        \label{fig:fashion_original}
    }
    \subfloat[CIFAR10]{
        \includegraphics[width=0.17\textwidth]{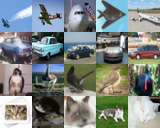}
        \label{fig:cifar_original}
    }
    \subfloat[TaxiNet]{
        \includegraphics[width=0.33\textwidth]{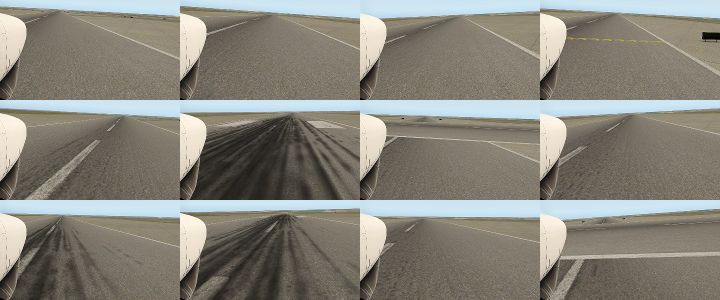}
        \label{fig:taxinet_original}
    }
    
    \subfloat[MNIST]{
        \includegraphics[width=0.17\textwidth]{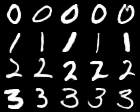}
        \label{fig:mnist_recon}
    }
    \subfloat[Fashion MNIST]{
        \includegraphics[width=0.17\textwidth]{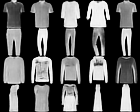}
        \label{fig:fashion_recon}
    }
    \subfloat[CIFAR10]{
        \includegraphics[width=0.17\textwidth]{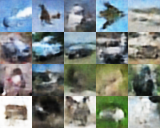}
        \label{fig:cifar_recon}
    }
    \subfloat[TaxiNet]{
        \includegraphics[width=0.33\textwidth]{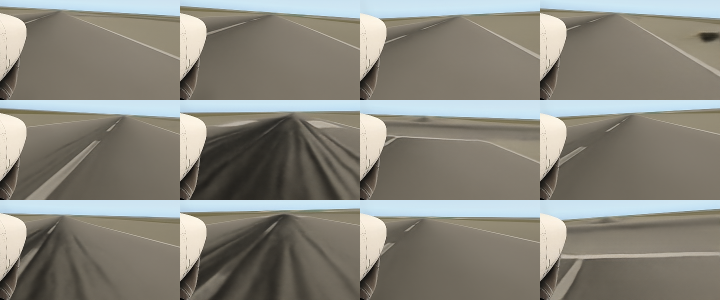}
        \label{fig:taxinet_recon}
    }
    \caption{Images reconstructed by trained VAEs---original images are in the upper row, reconstructed images are in the lower row. The reconstructed images look realistic overall, although some fine-grained texture is lost for Fashion MNIST and TaxiNet. CIFAR10 images, however, are too blurry that the objects are not always recongnizable.}
    \label{fig:reconstruction}
\end{figure*}

We used Fr\'echet Inception Distance (FID) as a quantitative measure of realism, and presented the scores of the generated fault-revealing tests in Table~\ref{tbl:result}.
The values indicate the degree of discrepancy between the original dataset and the generated one. A non-zero FID score indicates how different the generated inputs are different from the original dataset.
However, we observed that the score alone cannot be used as an objective measure to determine the realism, especially when comparing across different datasets---i.e. a dataset with a higher FID score may {\it look more real} than another dataset with a lower score.
As a complementary qualitative measure of visual quality, we also present the original and the reconstructed images side by side in Figure~\ref{fig:reconstruction}.
The reconstruction quality serves as a sanity check of trained VAEs---if the pair of encoder and decoder is well trained, it should at least be able to reconstruct the original validation dataset well.

\begin{figure*}
    \centering
    \subfloat[MNIST]{
        \includegraphics[width=0.23\textwidth]{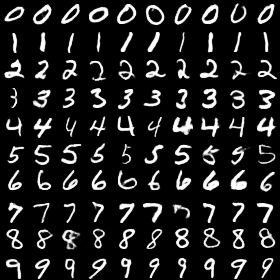}
        \label{fig:mnist_samples}
    }
    \subfloat[Fashion MNIST]{
        \includegraphics[width=0.23\textwidth]{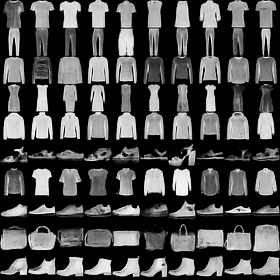}
        \label{fig:fashion_samples}
    }
    \subfloat[TaxiNet]{
        \includegraphics[width=0.418\textwidth]{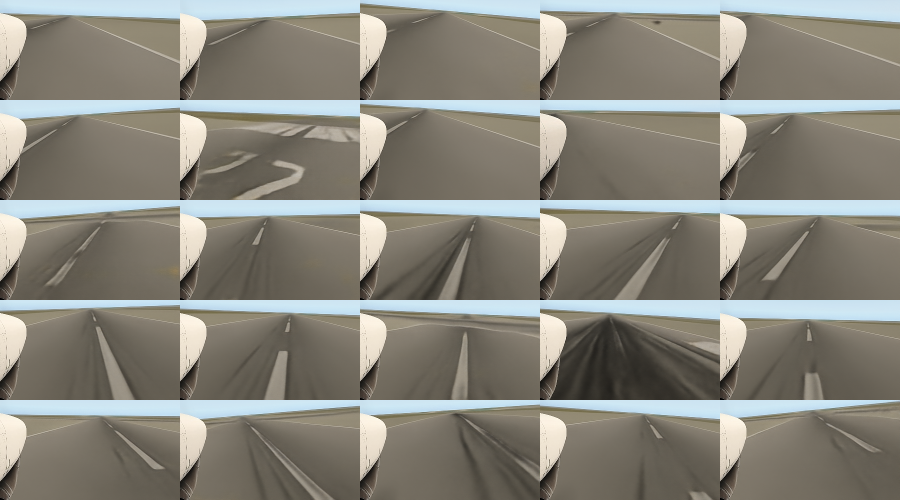}
        \label{fig:taxinet_samples}
    }
    \caption{
    Images generated by the Two-stage VAEs, sampled randomly from the second-stage manifold space. Images in each row are conditioned by the same class label (no cherry-picking).}
    \label{fig:vae_samples}
\end{figure*}

Figure~\ref{fig:vae_samples} shows the visual quality of the images generated by the trained VAEs using random sampling.
Each row of images are generated with the same class conditioning. From the top to the bottom, MNIST classes range from 0 to 9; Fashion MNIST classes are t-shirt/top, trouser, pullover, dress, coat, sandal, shirt, sneaker, bag, ankle boot; TaxiNet classes are far right, right, center, left, far left.
We can see that most of the images are very crisp and almost indistinguishable from the original images.
Each image matches the label it was conditioned with, and different varieties appear within the same class.

\subsection{Can we generate fault-revealing test cases?}

For each model under test, we attempted generating one thousand test cases.
We call it an {\it attempt} because some generated inputs may be invalid with respect to the real input distribution.
As there is a tension between the two optimization terms in Equation~\ref{eq:opt}, some inputs may be sampled from out of the distribution.
These out-of-distribution cases cannot be filtered out automatically because the real data distribution is beyond a logical specification.
We used human as oracle to determine whether each test case---or a pair of image and label---is valid or not.
We argue that this yes-or-no question is a much simpler task than labeling the image from the scratch, especially as the number of classes increases.

\begin{figure}[h]
    \centering
    \subfloat{
        \includegraphics[width=0.95\columnwidth]{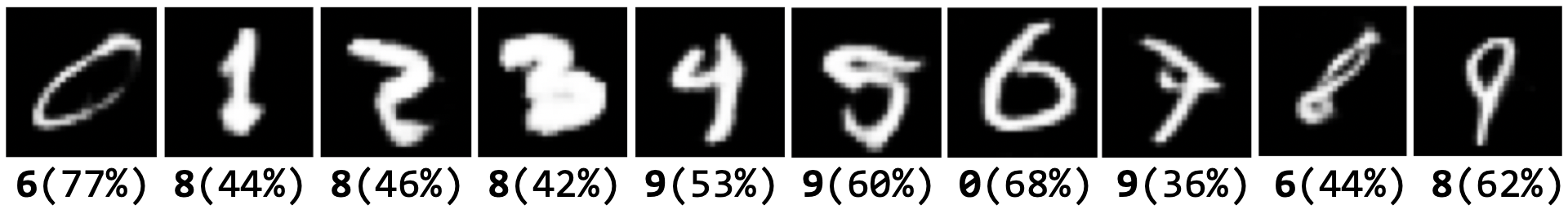}
        \label{fig:faults_mnist}
    }
    \hfill
    \centering
    \subfloat{
        \includegraphics[width=0.95\columnwidth]{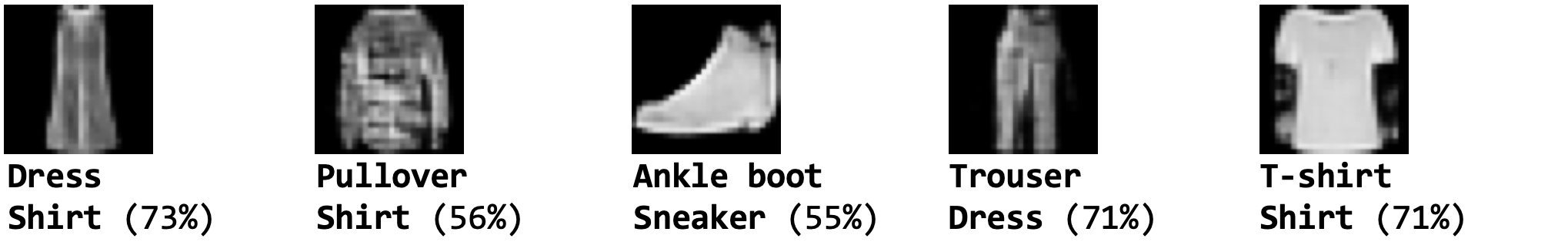}
        \label{fig:faults_fashion}
    }
    \caption{fault-revealing test cases for MNIST and Fashion MNIST (cherry-picked). The lower caption denote the prediction of the model under test, with the predicted probability marked inside the round brackets.}
    \label{fig:faults}
\end{figure}

Table~\ref{tbl:result} shows the statistics on generating fault-finding inputs.
For both MNIST and Fashion MNIST, with both random sampling and search-based optimization, we could generate hundreds of valid fault-revealing inputs in a reasonable amount of time.
Figure~\ref{fig:faults} showcases some of the fault-revealing test cases after filtering out invalid ones.
It can be seen that they are indeed corner-case inputs, yet still looking realistic.

\begin{table*}[]
    \centering
    \caption{Fault-finding test cases generated using the proposed technique. One thousand fault-finding cases were sampled by search-based approach and random sampling. Out-of-distribution images or cases with invalid labels were manually discarded.}
    \label{tbl:result}
    \input{tables/result.tex}
\end{table*}



\subsection{Is search better than random sampling?}

Table~\ref{tbl:result} compares the two methods side by side with the same goal of generating one thousand fault-revealing test cases.
Search-based optimization required significantly smaller number of test inputs to be generated in the first place before determining whether each input is fault-revealing.
However, this came with an added cost of measuring uncertainty of the model, which is far more expensive per each computation than simply computing the output and checking whether $y \ne \hat{y}$.
Thus, in terms of the cost, random sampling turned out to be much more effective than systematic search.

Another criterion for comparison is the ratio of valid inputs.
Although we added a term to favor sampling a more likely input, optimization towards higher uncertainty lead the generated images to be less realistic, as shown in the lower number of valid test cases with search-based method compared to random sampling.
In conclusion, at least with the given fitness function, random sampling from the latent space turned out to be much more effective and efficient than searching.

%% file: tables/result.tex
\centering
\footnotesize
\begin{tabular}{lrrrrrrrrrrr} 
\toprule
        & \multicolumn{5}{c}{Search} & \multicolumn{5}{c}{Random Sampling} & \multicolumn{1}{c}{\multirow{2}{*}{\begin{tabular}[c]{@{}c@{}}labeling\\cost \end{tabular}}}  \\ 
\cmidrule(lr){2-6}\cmidrule(lr){7-11}
        & \multicolumn{1}{l}{Generated} & \multicolumn{1}{l}{Time} & \multicolumn{1}{c}{FF. Count} & \multicolumn{1}{c}{Valid} & \multicolumn{1}{c}{FID} & \multicolumn{1}{l}{Generated} & \multicolumn{1}{l}{Time} & \multicolumn{1}{c}{FF. Count} & \multicolumn{1}{c}{Valid} & \multicolumn{1}{c}{FID} & \multicolumn{1}{c}{}  \\ 
\midrule
MNIST   & 3,324 & 17m 50s & 1000 & 310 & 31.11 $\pm$ 0.04 & 300,256 & 5m 42s  & 1000  & 663  & 29.36 $\pm$ 0.06 & 1.7s/ea. \\
Fashion & 3,617 & 23m 13s & 1000 & 542 & 66.90 $\pm$ 0.09 & 12,128 & 6m 57s & 1000 & 687 & 69.74 $\pm$ 0.19 & 2.0s/ea. \\
\bottomrule
\end{tabular}

%% file: sections/related.tex
A wide adoption of machine learning (ML) in the real-world applications is posing an unprecedented challenge to software testing due to the unique characteristics of the ML systems.
Many ML testing approaches had been suggested in recent years for examining properties such as the (adversarial) robustness~\cite{madry2017towards}, correctness, and fairness, among many.
For a comprehensive survey on machine learning testing, see the work by Zhang~\etal~\cite{zhang2019machine}.
Per their taxonomy, our work presents a novel test case generation technique---both input and oracle---with the goal of testing empirical correctness.

Since the discovery that image classification deep neural networks are easily fooled~\cite{nguyen2014deep,moosavi2016deepfool}, several attempts had been made to facilitate the testing process by generating test inputs~\cite{pei2017deepxplore,tian2017deeptest,zhang2018deeproad,zhou2018deepbillboard}.
Most of the approaches can be summarized as utilizing metamorphic testing~\cite{zhu2019datamorphic} in one way or another, and attempt image-to-image transformations from existing datapoints to synthesize new images.
DeepXplore~\cite{pei2017deepxplore} is one of the earliest pioneering work which proposed a white-box differential testing approach for generating new inputs.
They synthesized new images by applying constraint-preserving transformations on existing datapoints with the objective of jointly maximizing the neuron coverage and the disagreement among the committee of models under test.
Their generated inputs, however, are arguably {\it synthetic}, since the transformations---such as occlusions---are visually noticeable and seem unlikely to appear in the real world.
Synthetic test inputs---and any technique for adversarial input generation in the same regard---may be desirable if an adversary is assumed, but if the concern is system's correctness under normal operating condition, unrealistic inputs are merely false alarms, for they violate the environmental assumptions.

DeepTest~\cite{tian2017deeptest}, on the other hand, aimed at generating realistic images.
They applied metamorphic testing on autonomous driving application and showed that the domain-specific transformations such as fog, rain, and high contrast, can be used to reveal faults in the model under test.
DeepRoad~\cite{zhang2018deeproad} stretched this idea further for synthesizing realistic inputs in a sub-domain of the dataset.
They aimed at learning complex domain-specific metamorphic transformations so that new inputs can be synthesized beyond simple computer-vision-based transformations.
They used a variant of generative adversarial network called UNIT~\cite{liu2017unsupervised} for this transformation, which allowed them to transform sunny road images to snowy or rainy road images.
This approach, however, requires effort in constructing domain-specific image translators which inevitably involves collecting many images from which to learn the desired {\it style}.
In comparison, our approach does not require additional data beyond training dataset nor does it rely on metamorphic data transformation, yet is able to generate inputs beyond the training dataset.
Given these contrasting characteristics, we believe that the two approaches can complement each other.
Metamorphic testing can extrapolate the given dataset based on a set of rules, and our approach can interpolate thoroughly within the given dataset.


Yoo~\cite{yoo2019sbst} presented an idea of using search-based approaches for machine learning testing.
Our work presents a concretization of this idea, and answers the question with full-bodied development of the technique.

%% file: sections/conclusion.tex
This paper proposed a manifold-based test generation framework for automatically generating realistic yet fault-revealing test cases for testing machine-learning-enabled image classification models.
Built on top of Two-stage CVAE, the proposed method is capable of generating high-dimensional images and their associated labels automatically.
The experiments demonstrated that our approach can generate hundreds of fault-revealing test cases in a few minutes, and can reveal faults in the models that achieve high accuracy.

The idea proposed in this paper needs to be developed further and scrutinized more extensively.
Other fitness function can be designed and experimented for obtaining test cases with more desirable characteristics, such as being in-distribution.
Since fault-revealing inputs can be mapped to certain regions of the manifold space, an understanding on this space may lead to understanding the weakness of the model.
In future work, we also plan to extend our approach to regression tasks, where assigning labels can be even more expensive.

%% file: vaetestgen2.bbl
\begin{thebibliography}{10}
\providecommand{\url}[1]{#1}
\csname url@samestyle\endcsname
\providecommand{\newblock}{\relax}
\providecommand{\bibinfo}[2]{#2}
\providecommand{\BIBentrySTDinterwordspacing}{\spaceskip=0pt\relax}
\providecommand{\BIBentryALTinterwordstretchfactor}{4}
\providecommand{\BIBentryALTinterwordspacing}{\spaceskip=\fontdimen2\font plus
\BIBentryALTinterwordstretchfactor\fontdimen3\font minus
  \fontdimen4\font\relax}
\providecommand{\BIBforeignlanguage}[2]{{%
\expandafter\ifx\csname l@#1\endcsname\relax
\typeout{** WARNING: IEEEtran.bst: No hyphenation pattern has been}%
\typeout{** loaded for the language `#1'. Using the pattern for}%
\typeout{** the default language instead.}%
\else
\language=\csname l@#1\endcsname
\fi
#2}}
\providecommand{\BIBdecl}{\relax}
\BIBdecl

\bibitem{zhang2019machine}
J.~M. Zhang, M.~Harman, L.~Ma, and Y.~Liu, ``Machine learning testing: Survey,
  landscapes and horizons,'' \emph{CoRR}, vol. abs/1906.10742, 2019.

\bibitem{zhu2019datamorphic}
H.~{Zhu}, D.~{Liu}, I.~{Bayley}, R.~{Harrison}, and F.~{Cuzzolin},
  ``Datamorphic testing: A method for testing intelligent applications,'' in
  \emph{2019 IEEE International Conference On Artificial Intelligence Testing
  (AITest)}, April 2019, pp. 149--156.

\bibitem{tian2017deeptest}
Y.~Tian, K.~Pei, S.~Jana, and B.~Ray, ``Deeptest: Automated testing of
  deep-neural-network-driven autonomous cars,'' in \emph{Proceedings of the
  40th International Conference on Software Engineering}, ser. ICSE '18.\hskip
  1em plus 0.5em minus 0.4em\relax New York, NY, USA: ACM, 2018, pp. 303--314.

\bibitem{zhang2018deeproad}
M.~Zhang, Y.~Zhang, L.~Zhang, C.~Liu, and S.~Khurshid, ``{DeepRoad}: Gan-based
  metamorphic testing and input validation framework for autonomous driving
  systems,'' ser. ASE 2018.\hskip 1em plus 0.5em minus 0.4em\relax ACM, 2018.

\bibitem{berthelot2018understanding}
D.~Berthelot, C.~Raffel, A.~Roy, and I.~Goodfellow, ``Understanding and
  improving interpolation in autoencoders via an adversarial regularizer,''
  \emph{arXiv preprint arXiv:1807.07543}, 2018.

\bibitem{mcminn2011search}
P.~{McMinn}, ``Search-based software testing: Past, present and future,'' in
  \emph{2011 IEEE Fourth International Conference on Software Testing,
  Verification and Validation Workshops}, March 2011, pp. 153--163.

\bibitem{byun2019input}
T.~{Byun}, V.~{Sharma}, A.~{Vijayakumar}, S.~{Rayadurgam}, and D.~{Cofer},
  ``Input prioritization for testing neural networks,'' in \emph{2019 AITest},
  April 2019, pp. 63--70.

\bibitem{bengio2013representation}
Y.~{Bengio}, A.~{Courville}, and P.~{Vincent}, ``Representation learning: A
  review and new perspectives,'' \emph{IEEE Transactions on Pattern Analysis
  and Machine Intelligence}, vol.~35, no.~8, pp. 1798--1828, Aug 2013.

\bibitem{cayton2005algorithms}
L.~Cayton, ``Algorithms for manifold learning,'' \emph{Univ. of California at
  San Diego Tech. Rep}, vol.~12, no. 1-17, p.~1, 2005.

\bibitem{dai2019diagnosing}
B.~Dai and D.~P. Wipf, ``Diagnosing and enhancing {VAE} models,'' \emph{CoRR},
  vol. abs/1903.05789, 2019.

\bibitem{doersch2016tutorial}
C.~Doersch, ``Tutorial on variational autoencoders,'' \emph{ArXiv}, 2016.

\bibitem{larsen2015autoencoding}
A.~B.~L. Larsen, S.~K. S{\o}nderby, and O.~Winther, ``Autoencoding beyond
  pixels using a learned similarity metric,'' \emph{CoRR}, vol. abs/1512.09300,
  2015.

\bibitem{sohn2015learning}
K.~Sohn, H.~Lee, and X.~Yan, ``Learning structured output representation using
  deep conditional generative models,'' in \emph{NIPS}, 2015.

\bibitem{pei2017deepxplore}
K.~Pei, Y.~Cao, J.~Yang, and S.~Jana, ``{DeepXplore: Automated Whitebox Testing
  of Deep Learning Systems},'' in \emph{Proceedings of the 26th Symposium on
  Operating Systems Principles - SOSP '17}.\hskip 1em plus 0.5em minus
  0.4em\relax New York, New York, USA: ACM Press, 2017, pp. 1--18.

\bibitem{gal2016dropout}
Y.~Gal and Z.~Ghahramani, ``Dropout as a bayesian approximation: Representing
  model uncertainty in deep learning,'' ser. ICML'16.\hskip 1em plus 0.5em
  minus 0.4em\relax JMLR.org, 2016, pp. 1050--1059.

\bibitem{kennedy2010particle}
\BIBentryALTinterwordspacing
J.~Kennedy, \emph{Particle Swarm Optimization}.\hskip 1em plus 0.5em minus
  0.4em\relax Boston, MA: Springer US, 2010, pp. 760--766. [Online]. Available:
  \url{https://doi.org/10.1007/978-0-387-30164-8_630}
\BIBentrySTDinterwordspacing

\bibitem{lecun1998mnist}
Y.~LeCun, ``{The MNIST database of handwritten digits},''
  \emph{http://yann.lecun.com/exdb/mnist/}, 1998.

\bibitem{xiao2017fashion}
H.~Xiao, K.~Rasul, and R.~Vollgraf. (2017) Fashion-mnist: a novel image dataset
  for benchmarking machine learning algorithms.

\bibitem{krizhevsky2009learning}
A.~Krizhevsky, G.~Hinton \emph{et~al.}, ``Learning multiple layers of features
  from tiny images,'' Citeseer, Tech. Rep., 2009.

\bibitem{cai2018proxylessnas}
H.~Cai, L.~Zhu, and S.~Han, ``Proxylessnas: Direct neural architecture search
  on target task and hardware,'' \emph{CoRR}, vol. abs/1812.00332, 2018.

\bibitem{twostagevae}
\BIBentryALTinterwordspacing
B.~Dai and D.~P. Wipf. (2019) {TwoStageVAE}. [Online]. Available:
  \url{https://github.com/daib13/TwoStageVAE/}
\BIBentrySTDinterwordspacing

\bibitem{heusel2017gans}
M.~Heusel, H.~Ramsauer, T.~Unterthiner, B.~Nessler, and S.~Hochreiter, ``{GANs}
  trained by a two time-scale update rule converge to a local nash
  equilibrium,'' in \emph{NIPS 30}, 2017, pp. 6626--6637.

\bibitem{szegedy2015going}
C.~Szegedy, W.~Liu, Y.~Jia, P.~Sermanet, S.~Reed, D.~Anguelov, D.~Erhan,
  V.~Vanhoucke, and A.~Rabinovich, ``Going deeper with convolutions,'' in
  \emph{CVPR}, 2015, pp. 1--9.

\bibitem{madry2017towards}
A.~Madry, A.~Makelov, L.~Schmidt, D.~Tsipras, and A.~Vladu, ``Towards deep
  learning models resistant to adversarial attacks,'' \emph{arXiv preprint
  arXiv:1706.06083}, 2017.

\bibitem{nguyen2014deep}
A.~{Nguyen}, J.~{Yosinski}, and J.~{Clune}, ``Deep neural networks are easily
  fooled: High confidence predictions for unrecognizable images,'' in
  \emph{CVPR}, June 2015, pp. 427--436.

\bibitem{moosavi2016deepfool}
S.-M. Moosavi-Dezfooli, A.~Fawzi, and P.~Frossard, ``{DeepFool: a simple and
  accurate method to fool deep neural networks},'' in \emph{CVPR}, 2016.

\bibitem{zhou2018deepbillboard}
H.~Zhou, W.~Li, Y.~Zhu, Y.~Zhang, B.~Yu, L.~Zhang, and C.~Liu, ``Deepbillboard:
  Systematic physical-world testing of autonomous driving systems,''
  \emph{CoRR}, vol. abs/1812.10812, 2018.

\bibitem{liu2017unsupervised}
M.-Y. Liu, T.~Breuel, and J.~Kautz, ``Unsupervised image-to-image translation
  networks,'' in \emph{NIPS}, 2017.

\bibitem{yoo2019sbst}
S.~Yoo, ``{SBST} in the age of machine learning systems: challenges ahead,'' in
  \emph{Proceedings of the 12th International Workshop on Search-Based Software
  Testing}.\hskip 1em plus 0.5em minus 0.4em\relax IEEE Press, 2019, pp. 2--2.

\end{thebibliography}
